\documentclass{article}
\usepackage[utf8]{inputenc}
\usepackage{rotating}
\usepackage{multirow}
\usepackage{textcomp}
\newcommand{\pkg}[1]{#1}
\newcommand{\proglang}[1]{#1}
\newcommand{\code}[1]{\texttt{#1}}
\usepackage{Sweave}

\author{Miron Bartosz Kursa\\Interdisciplinary Centre for Mathematical and Computational Modelling\\University of Warsaw}
\title{\pkg{rFerns} -- Random Ferns Method Implementation for the General-Purpose Machine Learning}

\newcommand{\argmax}{\arg\max}

\newcommand{\ti}[1]{\textsc{#1}}
\newcommand{\lang}[1]{\proglang{#1}}
\newcommand{\mrv}[2]{\multirow{#1}{*}{\begin{sideways}#2\end{sideways}}}

\begin{document}
\maketitle

\abstract{
Random ferns is a very simple yet powerful classification method originally introduced for specific computer vision tasks.
In this paper, I show that this algorithm may be considered as a constrained decision tree ensemble and use this interpretation to introduce a series of modifications allowing one to use Random ferns in a general machine learning problems.
Moreover, I extend the method with internal error approximation and attribute importance measure based on a corresponding features of the Random forest algorithm.

I also present the \proglang{R} package \pkg{rFerns} containing an efficient implementation of such modified version of Random ferns.
}

\section{Introduction}
Random ferns is a machine learning algorithm proposed by \cite{Ozuysal2007} for matching same elements between two images of the same scene, allowing one to recognise certain objects or trace them on videos. 
The original motivation behind this method was to create a simple and efficient algorithm by extending the Naïve Bayes classifier; still the authors acknowledged its strong connection to the decision tree ensembles like the Random forest \cite{Breiman2001} algorithm.

Since introduction, Random ferns have been applied in numerous computer vision application, like image recognition \cite{Bosch2007}, action recognition \cite{Oshin2009} or augmented reality \cite{Wagner2010}.
However, it has not gathered attention outside this field; thus, this work aims to bring this algorithm to a much wider spectrum of applications.
In order to do that, I propose a generalised version of the algorithm, implemented as an \proglang{R} \cite{R} package \pkg{rFerns}.

The paper is organised as follows.
Section~\ref{sec:algo} briefly recalls the Bayesian derivation of the original version of Random ferns, presents the decision tree ensemble interpretation of the algorithm and lists modifications leading to the \pkg{rFerns} variant.
Next, in the Section~\ref{sec:pack}, I present the \pkg{rFerns} package and discuss the Random ferns incarnation of a two important features of the Random forest, internal error approximation and attribute importance measure.
Section~\ref{sec:bench} contains the assessment of \pkg{rFerns} in a several well known machine learning problems.
The results and computational performance of the algorithm are compared with Random forest implementation contained in the \pkg{randomForest} package \cite{Liaw2002}.
The paper is concluded in the Section~\ref{sec:conc}.

\section{Random ferns algorithm}
\label{sec:algo}
Following the original derivation, let's consider a classification problem based on an dataset $(X_{i,j},Y_i)$ with $p$ binary attributes $X_{\cdot,j}$ and $n$ objects $X_{i,\cdot}$ equally distributed over $C$ disjoint classes (those assumptions will be relaxed in the further part the paper).
The generic \textit{Maximum a Posteriori} (MAP) Bayes classifier classifies the object $X_{i,\cdot}$ as
\begin{equation}
Y^{p}_i=\argmax_y P(Y_i=y|X_{i,1},X_{i,2},\ldots,X_{i,p});
\end{equation}
according to the Bayes theorem, it is equal to
\begin{equation}
Y^{p}_i=\argmax_y P(X_{i,1},X_{i,2},\ldots,X_{i,p}|Y_i=y).
\end{equation}
Although this formula is strict, it is not practically usable due to a huge ($2^{p}$) number of possible $X_{i,\cdot}$ value combinations, most likely much larger than available number of training objects $n$ and thus making reliable estimation of probability impossible.

The simplest solution to this problem is to assume complete independence of the attributes, what brings us to the Naïve Bayes classification where
\begin{equation}
Y^{p}_i=\argmax_y \prod_j P(X_{i,j}|Y_i=y).
\end{equation}

The original Random ferns classifier \cite{Ozuysal2007} is an in-between solution defining a series of $K$ random selections of $D$ features ($\vec{j}_k\in \{1 .. P\}^D$, $k=1,\ldots,K$) treated using a corresponding series of simple exact classifiers (ferns), which predictions are assumed independent and thus combined in a naïve way, i.e.,
\begin{equation}
Y^{p}_i=\argmax_y \prod_k P(X_{i,\vec{j}_k}|Y_i=y),
\end{equation}
where $X_{i,\vec{j}_k}$ denotes $X_{i,j^{1}_{k}},X_{i,j^{2}_{k}},\ldots,X_{i,j^{D}_{k}}$.
This way one can still represent more complex interactions in the data, possibly achieving better accuracy than in purely naïve case.
On the other hand, such defined fern is still very simple and manageable for a range of $D$ values.

The training of the Random ferns classifier is performed through estimating probabilities $P(X_{i,\vec{j}_k}|Y_i=y)$ with empirical probabilities calculated from a training dataset $(X^t_{i,j},Y^t_i)$ of a size $n^t\times p$.
Namely, one uses frequencies of each class in each subspace of the attribute space defined by $\vec{j}_k$ assuming a Dirichlet prior, i.e.,
\begin{equation}
\widehat{P}(X_{i,\vec{j}_k}|Y_i=y)=\frac{1}{\#L_{i,\vec{j}_k}+C}\left(1+\#\left\{m\in L_{i,\vec{j}_k}:Y^t_m=y\right\}  \right),
\end{equation}
where $\#$ denotes the number of elements in a set and
\begin{equation}
L_{i,\vec{j}_k}=\left\{l\in \{1 .. n^t\}:\forall_{d\in 1..D} X^t_{l,j^d_k}=X_{i,j^d_k}\right\}
\label{eq:classLeaf}
\end{equation}
is the set of training objects in the same leaf of fern $k$ as object $i$.

\subsection{Ensemble of decision trees interpretation}

\begin{figure}[bt]
  \centering
  \includegraphics[width=\textwidth]{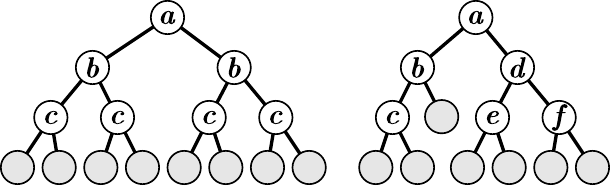}
  \caption{\label{fig:trefe} $D=3$ fern shown in a form of a decision tree (\textit{left}) and an example of a general decision tree (\textit{right}). Consecutive letters mark different splitting criteria.}
\end{figure}

A fern implements a partition of feature space into regions corresponding to all possible combinations of values of attributes $\vec{j}_k$.
This way it is equivalent to a binary decision tree of a depth $D$ for which all splitting criteria on a tree level $d$ are identical and split according to an attribute of index $j^d$, as shown on the Figure~\ref{fig:trefe}.
Consequently, because the attribute subsets $\vec{j}_k$ are generated randomly, the whole Random ferns classifier is equivalent to a random subspace \cite{Ho1998} ensemble of $K$ constrained decision trees.

Most ensemble classifiers combine predictions of its members through majority voting;
it is also the case for Random ferns when one consideres scores $S_{i,\vec{j}_k}(y)$ defined as
\begin{equation}
S_{i,\vec{j}_k}(y)=\log \widehat{P}(X_{i,\vec{j}_k}|Y_i=y)+\log C.
\end{equation}

This mapping effectively converts the MAP rule into majority voting
\begin{equation}
Y^{p}_i=\argmax_y \sum_k S_{i,\vec{j}_k}(y).
\end{equation}
Addition of $\log C$ causes that a fern that has no knowledge about the probability of classes for some object will give it a vector of scores equal zero.

\subsection{Introduction of bagging}

Using the ensemble of trees interpretation, in the \pkg{rFerns} implementation I was able to additionally combine random subspace with bagging, as it was shown to improve the accuracy of a similar ensemble classifiers \cite{Friedman2002a,Breiman2001,Panov2007}.
This method restricts training of each fern to \textit{bag}, a collection of objects selected randomly by sampling with replacement $n^t$ objects from an original training dataset, thus changing Equation~\ref{eq:classLeaf} into
\begin{equation}
L_{i,\vec{j}_k}=\left\{l\in B_k:\forall_{d\in 1..D} X^t_{l,j^d_k}=X_{i,j^d_k}\right\}
\end{equation}
where $B_k$ is a vector of indexes of the objects in the $k$-th fern's bag.

In such a set-up, the probability that a certain object won't be included in a bag is $(1-1/n^{t})^{n^{t}}$, thus each fern has a set of on average $n^t(1-1/n^t)^{n^t}$ ($n^{t}e^{-1}\approx{0.368}n^t$ for a large $n^t$) objects which were not used to build it.
They form \textit{out-of-bag} (OOB) subsets which will be denoted here as $B_k^\ast$.

\subsection{Generalisation beyond binary attributes}
As the original version of the Random ferns algorithm was formulated for datasets containing only binary attributes, the \pkg{rFerns} implementation had to introduce a way to also cope with continuous and categorical ones.
In the Bayesian classification view, this issue should be resolved by postulating and fitting some probability distribution over each attribute.
However, this approach introduces additional assumptions and possible problems connected to the reliability of fitting.

In the decision tree ensemble view, each non-terminal tree node maps certain attribute to a binary split using some criterion function, which is usually a greater-than comparison with some threshold value $\xi$ in case of continuous attributes (i.e., $f_\xi:x\rightarrow (x>\xi)$) and test whether it belongs to some subset of possible categories $\Xi$ in case of categorical attributes (i.e., $f_\Xi:x\rightarrow (x\in \Xi)$).

In most \textit{Classification And Regression Trees} (CART) and CART-based algorithms (including Random forest) the $\xi$ and $\Xi$ parameters of those functions are greedily optimised based on the training data to maximise the `effectiveness' of the split, usually measured by the information gain in decision it provides.
However, in order to retain the stochastic nature of Random ferns the \pkg{rFerns} implementation generates them at random, similar to the Extra-trees algorithm by \cite{Geurts2006}.
Namely, when a continuous attribute is selected for creation of a fern level a threshold $\xi$ is generated as a mean of two randomly selected values of it.
Correspondingly, for a categorical attribute $\Xi$ is set to a random one of all possible subsets of all categories of this attribute, except of two containing respectively all and none of the categories.

Obviously, completely random generation of splits can be less effective than optimising them in terms of the accuracy of a final classifier; the gains in computational efficiency may also by minor due to a fact that it does not change the complexity of the split building.
However, this way the classifier can escape certain overfitting scenarios and unveil more subtle interaction.
This and the more even usage of attributes may be beneficial both for the robustness of the model and the accuracy of the importance measure it provides.

While in this generalisation the scores depend on thresholds $\xi$ and $\Xi$, from now on I will denote them as $S_{i,F_k}$ where $F_k$ contains $\vec{j}_{k}$ and necessary thresholds.

\subsection{Unbalanced classes case}
When the distribution of the classes in the training decision vector becomes less uniform, its contribution to the final predictions of a Bayes classifier increases, biasing learning towards the recognition of larger classes.
Moreover, the imbalance may reach the point where it prevails the impact of attributes, making the whole classifier always vote on a largest class.

The original Random ferns algorithm was developed under assumption that the classes are equal, however such a case is very rare in a general machine learning and so the \pkg{rFerns} implementation has to cope with that problem as well.
Thus, it is internally enforcing balance of class' impacts by dividing the counts of objects of a certain class in a current leaf by the fraction of objects of that class in the bag of the current fern --- this is equivalent to a standard procedure of oversampling under-represented classes so that the amounts of objects of each class are equal within bag.

Obviously there exist exceptional use cases when such a heuristic may be undesired, for instance when the cost of misclassification is not uniform.
Then, it might be reversed or replaced with other prior by modifying the raw scores before the voting is applied.

\section[rFerns package]{\pkg{rFerns} package}
\label{sec:pack}

The training of a Random ferns model is performed by the \code{rFerns} function; it requires two parameters, the number of ferns $K$ and the depth of each one $D$, which should be passed via \code{ferns} and \code{depth} arguments respectively.
If not given, $K=1000$ and $D=5$ are assumed.
The current version of the package supports depths in range $1..15$.
The training set can be given either explicitly by passing predictor data frame and the decision vector, or via usual formula interface:
\begin{Schunk}
\begin{Sinput}
R> model <- rFerns(Species ~ ., data = iris, ferns = 1000, depth = 5)
R> model <- rFerns(iris[, -5], iris[, 5])
\end{Sinput}
\end{Schunk}
The results is a S3 object of a class \code{rFerns}, containing the ferns' structures $F_k$ and fitted scores' vectors for all leaves.

To classify new data, one should use the \code{predict} method of the \code{rFerns} class.
It will pull the dataset down each fern assigning each object with score vector from the leaf it ended in, sum the scores over the ensemble and finds the predicted classes.

For instance, let's set aside the even objects of \code{iris} data as a test set and train the model on the rest:
\begin{Schunk}
\begin{Sinput}
R> trainSet <- iris[c(TRUE, FALSE), ]
R> testSet <- iris[c(FALSE, TRUE), ]
R> model <- rFerns(Species ~ ., data = trainSet)
\end{Sinput}
\end{Schunk}
Then, the confusion matrix of predictions on a test set can be obtained by:
\begin{Schunk}
\begin{Sinput}
R> table(Prediction = predict(model, testSet), Truth = testSet[["Species"]])
\end{Sinput}
\begin{Soutput}
            Truth
Prediction   setosa versicolor virginica
  setosa         25          0         0
  versicolor      0         24         1
  virginica       0          1        24
\end{Soutput}
\end{Schunk}
Adding \code{scores=TRUE} to the \code{predict} call makes it return raw class scores.
The following code will extract scores of first three objects of each class in the test set:
\begin{Schunk}
\begin{Sinput}
R> testScores <- predict(model, testSet, scores = TRUE)
R> cbind(testScores, trueClass = testSet[["Species"]])[c(1:3, 26:28,
+     51:53), ]
\end{Sinput}
\begin{Soutput}
       setosa versicolor  virginica  trueClass
1   0.6083220 -0.8455781 -1.5057431     setosa
2   0.6672012 -0.9395135 -1.5468613     setosa
3   0.6255577 -0.9055999 -1.5120577     setosa
26 -1.3485694  0.2708711 -0.2595879 versicolor
27 -1.0709297  0.5555988 -0.7916823 versicolor
28 -1.2142952  0.4807975 -0.6461870 versicolor
51 -1.5922738 -0.1043981  0.3300516  virginica
52 -1.7083186 -0.2710267  0.4467801  virginica
53 -1.6010488 -0.6833165  0.6618151  virginica
\end{Soutput}
\end{Schunk}

\subsection{Error estimate}
By design, machine learning methods usually produce a highly biased results when tested on the training data; to this end, one needs to perform external validation to reliably assess its accuracy.
However, in a bagging ensemble we can perform a sort of internal cross-validation in which each train set object prediction is built by voting of only those of base classifiers which did not used this object for their training, i.e., which had it in their OOB subsets.
This idea has been originally used in the Random forest algorithm, and can be trivially transferred on any bagging ensemble, including \pkg{rFerns} version of Random ferns.
In this case the OOB predictions $Y^\ast_i$ will be given by
\begin{equation}
Y^\ast_i=\argmax_y \sum_{k:i\in B^\ast_k} S_{i,F_k}(y)
\end{equation}
and can be compared with the true classes $Y_i$ to calculate the OOB approximation of the overall error.

On the \proglang{R} level, OOB predictions are always calculated when training an \code{rFerns} model; when its corresponding object is printed, the overall OOB error and confusion matrix are shown, along with the training parameters:
\begin{Schunk}
\begin{Sinput}
R> print(model)
\end{Sinput}
\begin{Soutput}
 Forest of 1000 ferns of a depth 5.

 OOB error 5.33%; OOB confusion matrix:
            True
Predicted    setosa versicolor virginica
  setosa         25          0         0
  versicolor      0         24         3
  virginica       0          1        22
\end{Soutput}
\end{Schunk}
One can also access raw OOB predictions and scores by executing the \code{predict} method without providing new data to be classified:
\begin{Schunk}
\begin{Sinput}
R> oobPreds <- predict(model)
R> oobScores <- predict(model, scores = TRUE)
R> cbind(oobScores, oobClass = oobPreds, trueClass = trainSet$Species)[c(1:3,
+     26:28, 51:53), ]
\end{Sinput}
\begin{Soutput}
      setosa versicolor  virginica   oobClass  trueClass
1   277.9060 -407.55145 -602.21266     setosa     setosa
2   268.6165 -386.42968 -542.61370     setosa     setosa
3   304.4445 -451.73002 -648.26384     setosa     setosa
26 -425.2157   63.64338  -71.56243 versicolor versicolor
27 -573.6673   45.77139   24.85006 versicolor versicolor
28 -553.0338  113.32022  -49.42665 versicolor versicolor
51 -451.9256 -231.65799  207.49580  virginica  virginica
52 -571.6487 -222.92446  234.47927  virginica  virginica
53 -589.4160 -208.51861  229.84709  virginica  virginica
\end{Soutput}
\end{Schunk}

Note that for a very small values of $K$ some objects may manage to appear in every bag and thus get an undefined OOB prediction.

\subsection{Importance measure}
In addition to the error approximation, Random forest also uses the OOB objects to calculate the attribute importance.
It is defined as a difference in the accuracy on the original OOB subset and OOB subset with the values of a certain attribute permuted, averaged over all trees in the ensemble.

Such a measure can also be grafted on any bagging ensemble, including \pkg{rFerns}; moreover, one can make use of scores and replace the difference in accuracy with mean difference in score of the correct class, this way extracting importance information even from the OOB objects that are misclassified.
Precisely, such defined Random ferns importance of an attribute $a$ equals
\begin{equation}
I_a=\frac{1}{\#A(a)}\sum_{k\in A(a)}
\frac{1}{\#B_{k}^{\ast}}\sum_{i\in B_{k}^{\ast}}
\left (S_{i,F_k}(Y_i)- S^p_{i,F_k}(Y_i)\right),
\end{equation}
where $A(a)=\{k:a\in\vec{j}_k\}$ is a set of ferns that use attribute $a$ and $S^p_{i,F_k}$ is $S_{i,F_k}$ estimated on a permuted $X^{t}$ in which values of attribute $a$ have been shuffled.

One should also note that the fully stochastic nature of selecting attributes for building individual ferns guarantees that the attribute space is evenly sampled and thus all, even marginally relevant attributes are included in the model for a large enough ensemble.

Calculation of the variable importance can be triggered by adding \code{importance=TRUE} to the call to \code{rFerns}; then, the necessary calculations will be performed during the training process and the obtained importance scores placed into \code{importance} element of the \code{rFerns} object.
\begin{Schunk}
\begin{Sinput}
R> model <- rFerns(Species ~ ., data = iris, importance = TRUE)
R> model[["importance"]]
\end{Sinput}
\begin{Soutput}
             MeanScoreLoss SdScoreLoss
Sepal.Length     0.1748790 0.006270534
Sepal.Width      0.1578244 0.005121205
Petal.Length     0.3195912 0.010456676
Petal.Width      0.2796645 0.010555186
\end{Soutput}
\end{Schunk}

\section{Assessment}
\label{sec:bench}
I have tested \pkg{rFerns} on 7 classification problems from the \lang{R}'s \pkg{mlbench} \cite{Leish2010} package, namely DNA (\ti{dna}), Ionosphere (\ti{ion}), Pima Indian Diabetes (\ti{pim}), Satellite (\ti{sat}), Sonar (\ti{son}), Vehicle (\ti{veh}) and Vowel (\ti{vow}).

\subsection{Accuracy}
\begin{table}[h]
\begin{center}
\begin{tabular}{rr|cccc}
  \hline
&Set& \ti{dna} & \ti{ion} & \ti{pim} & \ti{sat} \\
 \hline
&Set size  & $3186\times 180$ & $351\times 34$ & $392\times 8$ & $6435\times 36$ \\
  \hline
\mrv{4}{OOB [\%]} & Ferns $5$ & $6.03  \pm 0.18$ & $7.32 \pm 0.23$ & $24.69 \pm 0.48$ & $18.40 \pm 0.13$ \\
& Ferns $10$   & $6.56  \pm 0.11$ & $7.35 \pm 0.22$ & $27.93 \pm 0.30$ & $15.46 \pm 0.06$ \\
& Ferns $D_b$  & $6.03 \pm 0.18$ & $7.07 \pm 0.40$ & $23.95 \pm 0.31$ & $14.33 \pm 0.05$ \\
& Forest & $4.13 \pm 0.09$ & $6.55 \pm 0.00$ & $21.76 \pm 0.36$ & $7.87 \pm 0.06$ \\
  \hline
\mrv{4}{CV [\%]} & Ferns  $5$  & $6.52 \pm 1.66$ & $7.78 \pm 3.41$ & $24.50 \pm 6.75$ & $18.60 \pm 1.32$ \\
&Ferns  $10$  & $6.96 \pm 1.30$ & $8.61 \pm 3.81$ & $29.50 \pm 6.10$ & $15.92 \pm 1.30$ \\
&Ferns  $D_b$  & $5.92 \pm 1.41$ & $5.00 \pm 3.88$ & $24.00 \pm 6.99$ & $14.32 \pm 0.88$ \\
& Forest & $4.20 \pm 0.99$ & $6.11 \pm 4.68$ & $21.00 \pm 3.94$ & $7.75 \pm 1.54$ \\
  \hline
&  $D_b$ & $5$ & $3$ & $7$ & $15$ \\
   \hline
 \hline
&Set& \ti{son} & \ti{veh} & \ti{vow} &  \\
  \hline
&Set size & $208\times 60$ & $846\times 18$ & $990 \times 10$ &  \\
\hline
\mrv{4}{OOB [\%]} &Ferns $5$ & $19.71 \pm 0.60$ & $31.17 \pm 0.49$ & $13.70 \pm 0.52$ &  \\
&Ferns $10$ & $14.18 \pm 1.12$ & $29.52 \pm 0.23$ & $4.42 \pm 0.26$ &  \\
&Ferns $D_b$ & $13.13 \pm 0.64$ & $28.83 \pm 0.49$ & $2.41 \pm 0.19$ &  \\
&Forest  & $15.38 \pm 0.64$ & $25.48 \pm 0.18$ & $2.13 \pm 0.11$ &  \\
  \hline
\mrv{4}{CV [\%]}&Ferns $5$ & $22.38 \pm 6.37$ & $32.94 \pm 4.15$ & $17.07 \pm 3.10$ &  \\
&Ferns $10$ & $14.29 \pm 5.94$ & $29.41 \pm 7.48$ & $5.25 \pm 1.64$ &  \\
&Ferns $D_b$ & $18.10 \pm 4.92$ & $28.71 \pm 5.69$ & $2.22 \pm 1.77$ &  \\
&Forest  & $19.52 \pm 8.53$ & $22.35 \pm 4.33$ & $2.22 \pm 1.70$ &  \\
  \hline
&$D_b$ & $12$ & $15$ & $15$ &  \\
   \hline

\end{tabular}

\caption{\label{tab:acc} OOB and cross-validation error of the Random ferns classifier for $5000$ ferns of a depth equal to $5$, $10$ and optimal over $\{1 .. 15\}$, $D_b$. Those results are compared to the accuracy of a Random forest classifier composed of $5000$ trees. Prediction errors are given as a mean and standard deviation over 10 repetitions of training for OOB and 10 iterations for cross-validation.}
\end{center}
\end{table}

For each of the testing sets, I have built $10$ Random ferns models for each of the depths in range $\{1 .. 15\}$ and number of ferns equal to $5000$ and collected the OOB error approximations.

Next, I have used those results to find optimal depths for each set ($D_b$) --- for simplicity I selected value for which the mean OOB error from all iterations was minimal.

Finally, I have verified the error approximation by running 10-fold stochastic cross-validation.
Namely, the set was randomly slit into test and training subsets, composed respectively of $10\%$ and $90\%$ of objects; the classifier was then trained on a training subset and its performance was assessed using the test set.
Such procedure has been repeated ten times.

As a comparison, I have also built and cross-validated $10$ Random forest models with $5000$ trees.
The ensemble size was selected so that both algorithm would manage to converge for all problems.

The results of those tests are collected in the Table~\ref{tab:acc}.
One can see that as in case of Random forest, OOB error approximation is a good estimate of the final classifier error.
It is also well serves as an optimisation target for the fern depth selection --- only in case of the Sonar data the naïve selection of the depth giving minimal OOB error led to a suboptimal final classifier, however one should note that the minimum was not significant in this case.

Based on the OOB approximations, forest outperforms ferns in all but one case; yet the results of cross-validation show that those differences are in practice masked by the natural variability of both classifiers.
Only in case of the Satellite data Random forest clearly achieves almost two times smaller error.

\subsection{Importance}
To test importance measure, I have used two sets for which importance of attributes should follow certain pattern.

\begin{figure}[ht]
  \centering
  \includegraphics[width=\textwidth]{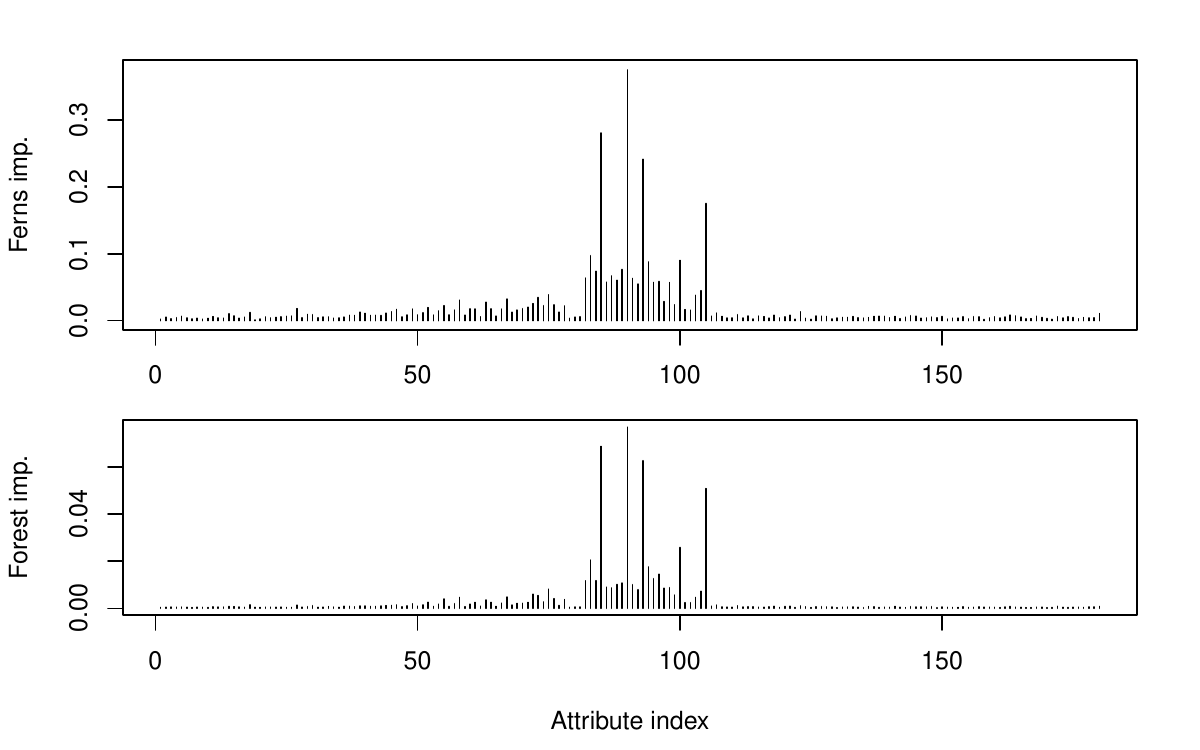}
  \caption{\label{fig:impDNA} Attribute importance for DNA data, generated by Random ferns (\textit{top}) and, for comparison, by Random forest (\textit{bottom}). Note that the importance peaks around 90th attribute, corresponding to an actual splicing site.}
\end{figure}
\begin{figure}[p]
  \centering
  \includegraphics[width=\textwidth]{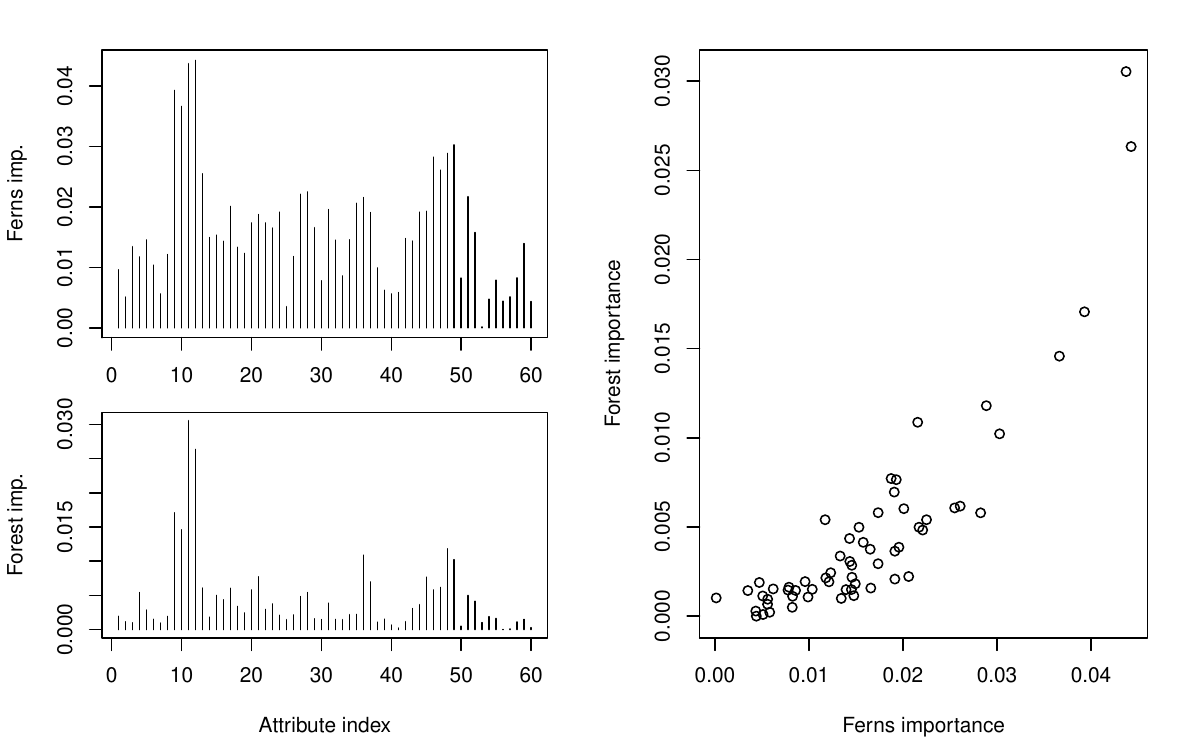}
  \caption{\label{fig:impSon} Importance measure for Sonar data, generated by Random ferns (\textit{top right}) and, for comparison, by Random forest (\textit{bottom right}). Those two measures are compared on a scatterplot (\textit{left}).}
\end{figure}
\begin{table}[p]
\begin{center}
\begin{tabular}{rr|c|cc|cc|c}
  \hline
& Set & Forest [s] & Ferns $10$ [s] & Speedup & Ferns $D_b$ [s] & Speedup & $D_b$ \\
  \hline
\mrv{7}{Just training}&  \ti{dna} & $30.13$ & $1.96$ & $15.41$ & $0.65$ & $46.28$ & $5$ \\
&  \ti{ion} & $3.41$ & $0.81$ & $4.21$ & $0.06$ & $55.35$ & $3$ \\
&  \ti{pim} & $2.45$ & $0.97$ & $2.53$ & $0.24$ & $10.29$ & $7$ \\
&  \ti{sat} & $136.91$ & $4.08$ & $33.55$ & $75.55$ & $1.81$ & $15$ \\
&  \ti{son} & $3.40$ & $0.78$ & $4.37$ & $2.95$ & $1.15$ & $12$ \\
&  \ti{veh} & $8.00$ & $1.67$ & $4.80$ & $46.04$ & $0.17$ & $15$ \\
&  \ti{vow} & $93.92$ & $4.77$ & $19.69$ & $140.26$ & $0.67$ & $15$ \\
  \hline
\mrv{7}{With importance}&\ti{dna} & $456.16$ & $4.37$ & $104.38$ & $1.87$ & $244.31$ & $5$ \\
&  \ti{ion} & $7.34$ & $1.46$ & $5.02$ & $0.25$ & $29.32$ & $3$ \\
&  \ti{pim} & $3.53$ & $1.06$ & $3.34$ & $0.35$ & $10.17$ & $7$ \\
&  \ti{sat} & $289.26$ & $8.77$ & $32.98$ & $82.80$ & $3.49$ & $15$ \\
&  \ti{son} & $4.83$ & $0.93$ & $5.18$ & $3.13$ & $1.54$ & $12$ \\
&  \ti{veh} & $17.26$ & $2.22$ & $7.77$ & $47.00$ & $0.37$ & $15$ \\
&  \ti{vow} & $99.26$ & $5.40$ & $18.39$ & $141.13$ & $0.70$ & $15$ \\
   \hline
\end{tabular}

\caption{\label{tab:tim} Training times of the \pkg{rFerns} and \pkg{randomForest} models made for $5000$ base classifiers, with and without importance calculation. Times are given as a mean over 10 repetitions.}
\end{center}
\end{table}

Each objects in the DNA set \cite{Noordewier1991} represent 60-residue DNA sequence in a way so that each consecutive triplet of attributes encodes one residue.
Some of the sequences contain a boundary between exon and intron (or intron and exon\footnote{The direction of a DNA sequence is significant, so those are separate classes.}) regions of the sequence --- the objective is to recognise and classify those sequences.
All sequences were aligned in a way that the boundary always lies between $30$th and $31$st residue; while the biological process of recognition is local, the most important attributes should be those describing residues in the vicinity of the boundary.

Objects in the Sonar set \cite{Gorman1988} correspond to echoes of a sonar signal bounced off either a rock or a metal cylinder (a model of a mine).
They are represented as power spectra, thus each next attribute value corresponds to the signal power contained within a consecutive frequency interval.
This way one may expect that there are frequency bands in which echoes significantly differ between classes, what would manifest as a set of peaks in the importance measure vector.

For both of this sets, I have calculated the importance measure using $1000$ ferns of a depth $10$.
As a baseline, I have used importance calculated using Random forest algorithm with $1000$ trees.

The results are presented on Figure~\ref{fig:impDNA} and Figure~\ref{fig:impSon}.
The importance measures obtained is both cases are consistent with the expectations based on the sets' structures --- for DNA, one can notice a maximum around attributes $90$--$96$, corresponding the actual cleavage site location.
For Sonar, the importance scores reveal a band structure which likely corresponds to the actual frequency intervals in which the echoes differ between stone and metal.

Both results are also qualitatively in agreement with those obtained from Random forest models.
Quantitative difference comes form the completely different formulations of both measures and possibly the higher sensitivity of ferns resulting from its fully stochastic construction.

\subsection{Computational performance}
In order to compare training times of \pkg{rFerns} and \pkg{randomForest} codes, I have trained both models on all $7$ benchmark sets for $5000$ ferns/trees, and, in case of ferns, depths $10$ and $D_b$.
Than I have repeated this procedure, this time making both algorithms calculate importance measure during training.

I have repeated both tests $10$ times to stabilise the results and collected the mean execution times; the results are collected in the Table~\ref{tab:tim}.
The results show that the usage of \pkg{rFerns} may result is significant speedups in certain applications; best speedups are achieved for the sets with larger number of objects, which is caused by the fact that Random ferns' training time scales linearly with the number of objects, while Random forest's $\sim n\log n$.

Also the importance can be calculated significantly faster by \pkg{rFerns} than by \pkg{randomForest}, and the gain increases with the size of the set.

\pkg{rFerns} is least effective for sets which require large depths of the fern --- in case of Vowel and Vehicle sets it was even slower than Random forest.
However, one should note that while the complexity of Random ferns $\sim 2^D$, its accuracy usually decreases much less dramatically when decreasing $D$ from its optimal value --- this way one may expect an effective trade-off between speed and accuracy.

\section{Conclusions}
\label{sec:conc}
In this paper, I have presented \pkg{rFerns}, a general-purpose implementation of the Random ferns, a fast, ensemble-based classification method.
Slight modifications of the original algorithm allowed me to additionally implement OOB error approximation and attribute importance measure.

Presented benchmarks showed that such algorithm can achieve accuracies comparable to Random forest algorithm while usually being much faster, especially for large datasets.

Also the importance measure proposed in this paper can be calculated very quickly and proved to behave in a desired way and be in agreement with the results of Random forest; however the in-depth assessment of its quality and usability for feature selection and similar problems requires further research.

\section*{Acknowledgements}
This work has been financed by the National Science Centre, grant 2011/01/N/ST6/07035.
Computations were performed at ICM, grant G48-6.

\bibliography{ferns}{}

\end{document}